\documentclass[10pt,letterpaper]{article}

\usepackage{cogsci}

\cogscifinalcopy % Uncomment this line for the final submission 
\usepackage{algorithm}
\usepackage{algorithmic}
\usepackage{amsmath,amssymb,amsfonts}
\usepackage{subfig}
\usepackage{booktabs}
\usepackage{setspace}
\usepackage{multirow}
\usepackage{graphicx}

\newenvironment{packeditemize}{\begin{list}{$\bullet$}{\setlength{\itemsep}{1pt}\addtolength{\labelwidth}{0pt}\setlength{\leftmargin}{\labelwidth}\setlength{\listparindent}{\parindent}\setlength{\parsep}{0pt}\setlength{\topsep}{0pt}}}{\end{list}}

\usepackage{pslatex}
\usepackage{apacite}
\usepackage{float} % Roger Levy added this and changed figure/table
\title{A Multimodal In Vitro Diagnostic Method for Parkinson’s Disease Combining Facial Expressions and Behavioral Gait Data}
 
\author{{\large \bf Wei~Huang\textsuperscript{1,2}},  
        {\large \bf Yinxuan~Xu\textsuperscript{1\dag}}, 
        {\large \bf Yintao~Zhou\textsuperscript{1\dag}},
        {\large \bf Zhengyu~Li\textsuperscript{3}},
        {\large \bf Jing~Huang\textsuperscript{3}},
        {\large \bf Meng~Pang\textsuperscript{1*}}\\
        {mengpang@ncu.edu.cn}\\
        {\textsuperscript{1}School of Mathematics and Computer Sciences, Nanchang University, Nanchang, China}\\
        {\textsuperscript{2}Yichun University, Yichun, China}\\
        {\textsuperscript{3}Nanchang University Second Affiliated Hospital, Nanchang, China}}

\begin{document}
\maketitle

\begin{abstract}
Parkinson's disease (PD), characterized by its incurable nature, rapid progression, and severe disability, poses significant challenges to the lives of patients and their families. Given the aging population, the need for early detection of PD is increasing. In vitro diagnosis has garnered attention due to its non-invasive nature and low cost. However, existing methods present several challenges: 1) limited training data for facial expression diagnosis; 2) specialized equipment and acquisition environments required for gait diagnosis, resulting in poor generalizability; 3) the risk of misdiagnosis or missed diagnosis when relying on a single modality. To address these issues, we propose a novel multimodal in vitro diagnostic method for PD, leveraging facial expressions and behavioral gait. Our method employs a lightweight deep learning model for feature extraction and fusion, aimed at improving diagnostic accuracy and facilitating deployment on mobile devices. Furthermore, we have established the largest multimodal PD dataset in collaboration with a hospital and conducted extensive experiments to validate the effectiveness of our proposed method.

\end{abstract}

\section{Introduction}

Parkinson's disease (PD) is a prevalent neurodegenerative disorder with inconspicuous early symptoms. By the time overt motor impairments manifest, the disease often progresses to an advanced stage, severely impacting patients' daily life. Recent data indicates that PD incidence has increased by 2.7 times over the past 30 years, affecting over 11.8 million people, and PD-related deaths have risen by 1.6 times to 388,000 \cite{steinmetz2024global}. While there is currently no effective prevention or treatment, medications like levodopa can alleviate symptoms and improve patients' quality of life. Therefore, early diagnosis and prognostic management are crucial for optimizing drug efficacy, slowing disease progression, and reducing the economic burden \cite{buono2021anxiety,gray2022long}.
\footnotetext{\textsuperscript{\dag}These authors contributed to the work equally.}
\footnotetext{\textsuperscript{*}Corresponding author.
}

\begin{figure}[t]
\setlength{\abovecaptionskip}{3mm}
\centering
\includegraphics[width=0.96\linewidth]{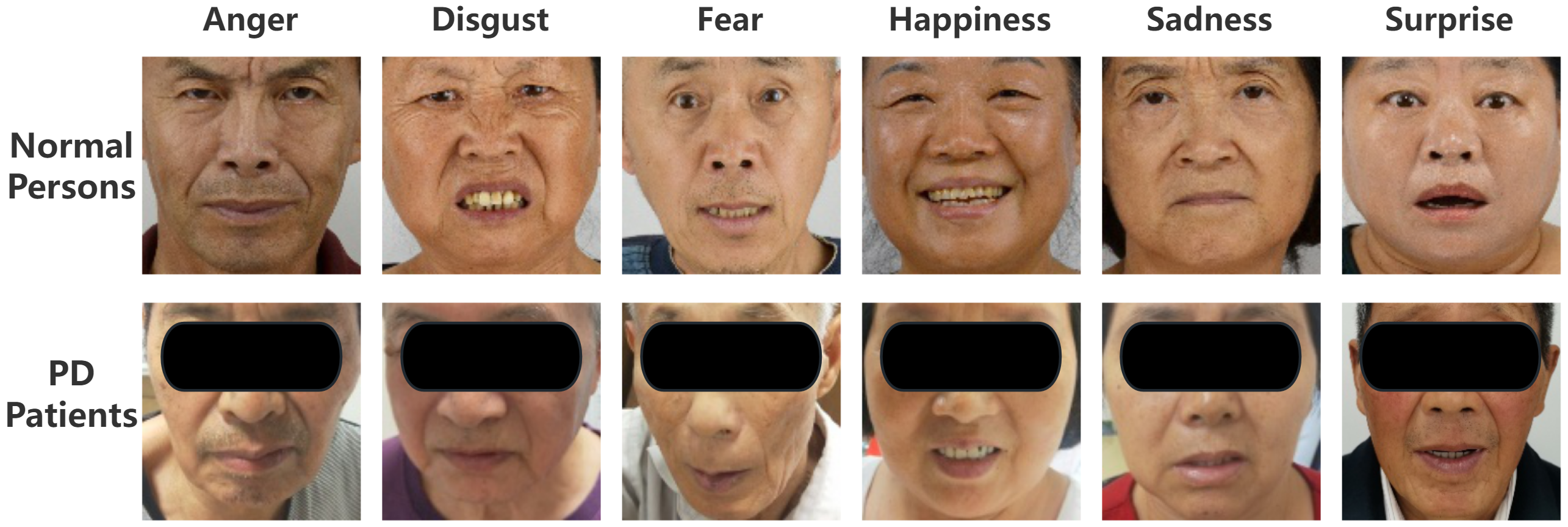}
\caption{A comparison between the masked faces of PD patients and the facial expression images of normal persons under six basic emotions (i.e., anger, disgust, fear, happiness, sadness, and surprise.)}
\label{FE-compare}
\vspace{-5mm}
\end{figure}

Generally, PD diagnosis can be broadly categorized into two types: in vivo and in vitro diagnosis. In vivo diagnosis primarily relies on specialized medical imaging techniques, such as CT, MRI, and PET \cite{tolosa2021challenges}. While these methods offer high accuracy, they also have limitations: they require professional equipments operated by trained personnel, and certain patients, like pregnant women or those with specific medical conditions, may not be suitable for some scans. Moreover, the high costs burden ordinary families. In contrast, in vitro diagnostic methods involve collecting biomarkers from PD patients, including voice signals, gait signals, and facial expressions, offering advantages such as being non-invasive, convenient, and having fewer patient restrictions. This makes them increasingly popular among doctors and researchers. Among these in vitro biomarkers, facial expressions and behavioral gait are particularly suited for routine diagnosis due to their universal applicability across races and languages, and the rich, stable visual feature information they provide.

Recent studies \cite{ricciardi2020hypomimia} have indicated that 70\% of PD patients exhibit facial expression dysfunction (commonly referred to as ``masked faces") when compared to non-PD patients, as illustrated in Figure. \ref{FE-compare}. While facial expression disorders have been utilized as a diagnostic criterion \cite{hou2022visual}, a small subset of patients in the early stages of the disease can still accurately express one or several categories of basic emotions. Consequently, relying solely on facial expressions for diagnosis can possibly result in missed diagnoses.
Furthermore, regarding behavioral gait diagnosis, the primary symptoms of PD include bradykinesia, myotonia, resting tremor, and abnormalities in posture and gait. Some researchers have conducted studies focusing on this biological signal \cite{liu2022quantitative}. %tosserams2022evaluation}
However, it is noteworthy that elderly PD patients often exhibit symptoms such as bradykinesia and postural instability, which can also be natural manifestations at this age. Therefore, relying on a single gait signal for PD diagnosis may lead to misdiagnoses.

\begin{figure*}[t]
\centering
\includegraphics[width=0.99\linewidth]{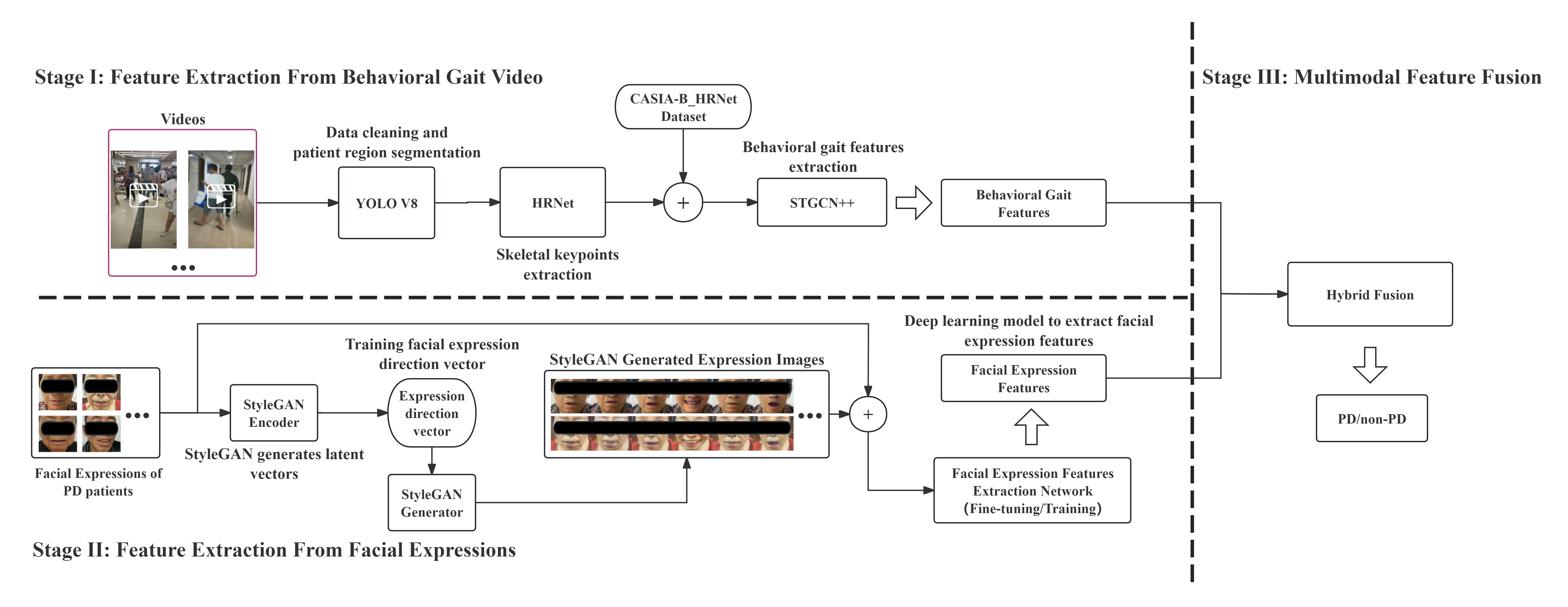}
\caption{Overview of the proposed multimodal in vitro diagnostic method for PD.}
\label{Flowchart}
\end{figure*}

To address the limitations of existing unimodal in vitro diagnostic methods for PD, we have collaborated with the affiliated hospital of Nanchang University to create the largest known PD multimodal dataset, PDMM. This dataset encompasses video recordings of seven facial expressions (neutral, anger, disgust, fear, happiness, sadness, surprise) and gait from 95 PD patients. Based on this dataset, we innovatively propose a multimodal in vitro PD diagnostic method that integrates facial expression and behavioral gait analysis. The flowchart of the proposed method is shown in Figure \ref{Flowchart}, which includes three core stages: Firstly, we preprocess the PDMM video data to segment the patient regions, utilize HRNet \cite{sun2019deep} to obtain skeletal keypoints, and then extract patients' behavioral gait features through STGCN++ \cite{duan2022pyskl}; Secondly, with the help of StyleGAN \cite{karras2020analyzing}, we generate facial expression images depicting other six basic emotions from a single neutral facial expression image of PD patients, thereby simulating their pre-morbid facial expression state as a reference group to train a discriminative model and extract highly discriminative facial expression features; Lastly, we propose a novel feature fusion strategy, namely hybrid fusion, aimed at effectively integrating the extracted behavioral gait and facial expression features, and based on the above process, we design an end-to-end multimodal data fusion and diagnostic model for PD prediction.

We summarize the contributions of this work as follows:

\begin{packeditemize}
\item We have created the PDMM dataset, which encompasses diverse facial expressions and gait video data from 95 PD patients. This dataset is currently the largest and most comprehensive multimodal dataset available for in vitro diagnostic research on PD.
\item We have pioneered a novel in vitro diagnostic method for PD, marking the first attempt to fuse multimodal data from facial expressions and behavioral gait. Specifically, we extracted skeletal key points from gait video data to obtain deep semantic features of PD patients' behavioral gait and used StyleGAN to generate premorbid expressions for training discriminative models and extracting highly discriminative facial expression features. 
\item We have introduced a hybrid feature fusion strategy to effectively integrate behavior gait and facial expression features for PD diagnosis. 
\item Our extensive experiments demonstrate the effectiveness of StyleGAN in generating virtual facial expression data of PD patients before symptom onset, as well as the superior performance of our proposed multimodal diagnostic method in PD diagnosis.
\end{packeditemize}

\section{RELATED WORKS}

 {\bf In vitro PD Diagnosis based on Behavioral Gait Signals.}
PD patients often exhibit bradykinesia, myotonia, and postural instability, which can impact gait patterns \cite{o2002dual}. Various studies have employed specialized equipment for quantitative gait analysis in PD diagnosis, such as piezoelectric sensors worn on the feet \cite{blin1990quantitative}, motion sensors \cite{havinga2007sensorshoe}, and Kinect sensors capturing 3D human skeleton motion \cite{li2018classification}. However, these methods are not suitable for large-scale screening due to limitations in space, hardware, cost, and operational complexity.
In contrast, gait video-based in vitro diagnosis of PD offers advantages such as easy operation, no need for wearable devices, low cost, no site limitations, and potential support from artificial intelligence. Researchers have analyzed PD patients using video data and various models, such as a 3D convolutional network model~\cite{yin2021assessment} and a two-stream spatio-temporal attention graph convolutional network for gait dyskinesia evaluation~\cite{guo2021multi}.

 {\bf In vitro PD Diagnosis based on Facial Expressions.}
In addition to speech and gait signals, recent research has explored facial expressions as a novel in vitro biomarker for diagnosing PD. This idea is supported by studies indicating that PD patients often exhibit inconspicuous facial expressions, described as ``masked faces" \cite{ricciardi2020hypomimia} %,bandini2017analysis}
. Vinokurov \emph{et al.} \cite{vinokurov2015quantifying} used 3D sensors for PD classification and diagnosis via linear regression, while Bo \emph{et al.} \cite{jin2020diagnosing} utilized the Face++ platform to extract facial expression features for diagnosis using an LSTM neural network. Huang \emph{et al.} \cite{huang2022facial,huang2023auto} achieved a good accuracy in PD diagnosis by extracting features from original and generated facial expressions of PD patients and classifying them using a deep neural network. 

It is worth noting that, the aforementioned PD diagnostic methods are all based on single-modal signals, and may lack strong robustness in real and complex clinical scenarios. For example, relying solely on facial expressions may lead to miss diagnoses, especially for early-stage patients who can still express emotions accurately. Additionally, elderly PD patients often exhibit symptoms like bradykinesia and postural instability, which can be age-related manifestations. Thus, relying on a single gait signal may lead to misdiagnoses. Given the lack of cross-linguistic universality in speech signals, this paper thus proposes a multi-modal PD diagnosis solution fusing gait and facial expressions, which aims to address the deficiencies of single-modal methods, as well as improve the PD diagnosis accuracy.

\section{THE PROPOSED METHOD}

%As shown in Figure \ref{Flowchart}, the proposed multimodal PD diagnostic method comprises three parts: Firstly, we segment patient regions from behavioral gait videos using the YOLOv8 model \cite{terven2023comprehensive}, extract skeletal keypoints with HRNet, and then extract behavioral gait features via STGCN++. Secondly, we use StyleGAN to generate facial expression images depicting six basic emotions from a single neutral facial image of PD patients, approximating their pre-morbid facial states. These images are used to train a deep learning model for extracting highly discriminative facial features. Lastly, we propose a feature fusion strategy, i.e., hybrid fusion, to effectively integrate extracted behavioral gait and facial expression features for PD diagnosis.

As shown in Figure \ref{Flowchart}, the proposed multimodal PD diagnostic method comprises three parts: Firstly, we employ the YOLOv8 model \cite{terven2023comprehensive} to segment patient areas from behavioral gait videos, subsequently extracting skeletal keypoints using HRNet, and further deriving behavioral gait features through STGCN++. Secondly, we use StyleGAN to generate facial expression images depicting six basic emotions from a single neutral facial image of PD patients, approximating their pre-morbid facial states. These images are used to train a deep learning model for extracting highly discriminative facial features. Lastly, we propose a feature fusion strategy, i.e., hybrid fusion, to effectively integrate extracted behavioral gait and facial expression features for PD diagnosis.

%we segment patient regions from behavioral gait videos using the YOLOv8 model \cite{terven2023comprehensive}, extract skeletal keypoints with HRNet, and then extract behavioral gait features via STGCN++. Secondly, we use StyleGAN to generate facial expression images depicting six basic emotions from a single neutral facial image of PD patients, approximating their pre-morbid facial states. These images are used to train a deep learning model for extracting highly discriminative facial features. Lastly, we propose a feature fusion strategy, i.e., hybrid fusion, to effectively integrate extracted behavioral gait and facial expression features for PD diagnosis.

%As illustrated in Figure \ref{Flowchart}, the multimodal PD diagnostic approach we propose encompasses three primary components: Initially, we employ the YOLOv8 model \cite{terven2023comprehensive} to segment patient areas from behavioral gait videos, subsequently extracting skeletal keypoints using HRNet, and further deriving behavioral gait features through STGCN++. In the second phase, we utilize StyleGAN to synthesize facial expression images representing six fundamental emotions from a single neutral facial image of PD patients, thereby estimating their pre-morbid facial configurations. These generated images serve as the basis for training a deep learning model to extract highly distinctive facial features. Finally, we introduce a feature fusion technique known as hybrid fusion, which efficiently integrates the extracted behavioral gait and facial expression features for the purpose of PD diagnosis.

\subsection{Feature Extraction From Behavioral Gait Video}

The behavioral gait feature extraction process necessitated preprocessing due to the less rigorous recording conditions and interfering factors present in the gait video data of PD patients. The process can be distilled into three key steps:
\begin{packeditemize}
    \item \textbf{Data cleaning and patient region segmentation:} We manually screened the behavioral gait videos to select unobstructed and clear clips, eliminating interfering factors such as the patients' rising and turning processes to obtain purer gait data. After cleaning, some videos contained multiple individuals. To eliminate interference from non-PD patients, we employed the YOLOv8 model for accurate tracking and segmentation, ensuring analysis focused solely on the region of interest (the PD patient).
    \item \textbf{Skeletal keypoints extraction:} After segmentation, we utilized HRNet to extract skeletal keypoints from the data, ensuring that these keypoints adhered to the COCO17 standard \cite{lin2014microsoft}.
    \item \textbf{Behavioral gait feature extraction:} We employed STGCN++, a modified version of the original STGCN model \cite{yan2018spatial}, to extract effective gait features from the collected skeletal keypoints. STGCN++ incorporates multi-branch time-domain convolution, enhancing temporal modeling capability while reducing computational complexity.
\end{packeditemize}

\subsection{Feature Extraction From Facial Expressions}

Due to the lack of control samples from PD patients before they developed the disease, it is challenging to train an effective discriminative model for feature extraction from the facial expression data we collected. To address this issue, we attempt to use StyleGAN to synthesize virtual facial expression images of PD patients in their pre-morbid state. Then, we explore the use of different deep learning models for feature extraction and classification training on the augmented facial expression data. The process for facial expression feature extraction from PD patients is detailed below.

\textbf{Step 1: Generating latent vectors using StyleGAN.} 
We employ a pre-trained StyleGAN generator combined with an encoder network, inspired by \cite{pang2024heterogeneous}. During training, we optimize the encoder parameters to ensure that the reconstructed images generated by the generator are as close as possible to the original input images. This approach enables us to obtain a latent vector in the latent space of StyleGAN that is similar to the identity information of the original input image. It should be noted that during training, the parameters of the encoder network are fine-tuned through a carefully designed similarity loss function, which is a weighted combination of the VGG-16 perceptual loss and the per-pixel mean squared error (MSE) loss. The definition of this loss function is as follows:
\begin{equation}
\label{eq1}
c^*=\min_{c}L_{percerp}(G(c),I_{O})+\frac{\lambda_{mse}} N\parallel G(c)-I_{O}\parallel_{2}^{2},
\end{equation}
where $c$ represents the target latent vector, $I_{O}$ denotes the input image, $G$ signifies the pre-trained StyleGAN generator, $N$ denotes the total number of image pixels, and $\lambda_{mse}$ is a weight hyperparameter. 
The perceptual loss, i.e., $L_{percep}$, measures the perceptual difference between the original image $I_O$ and the synthesized image $G(c)$ using a pre-trained VGG network. This perceptual loss is defined as 
\begin{equation}
\label{eq2}
L_{percep}(G(c),I_{O})=\sum_{j=1}^k\frac{\lambda_j}{N_j}\parallel C_j(I_O)-C_j(G(c))\parallel_2^2,
\end{equation}
where $C_j(.)$ denotes the feature map output from the $j$-th convolutional layer of the VGG-16 network, $k$ denotes the number of convolutional layers, $N_j$ represents the total number of pixels in the feature map from the $j$-th convolutional layer, and $\lambda_j$ is a weight hyperparameter.

\begin{figure}[t]
\centering
\includegraphics[width=0.8\linewidth]{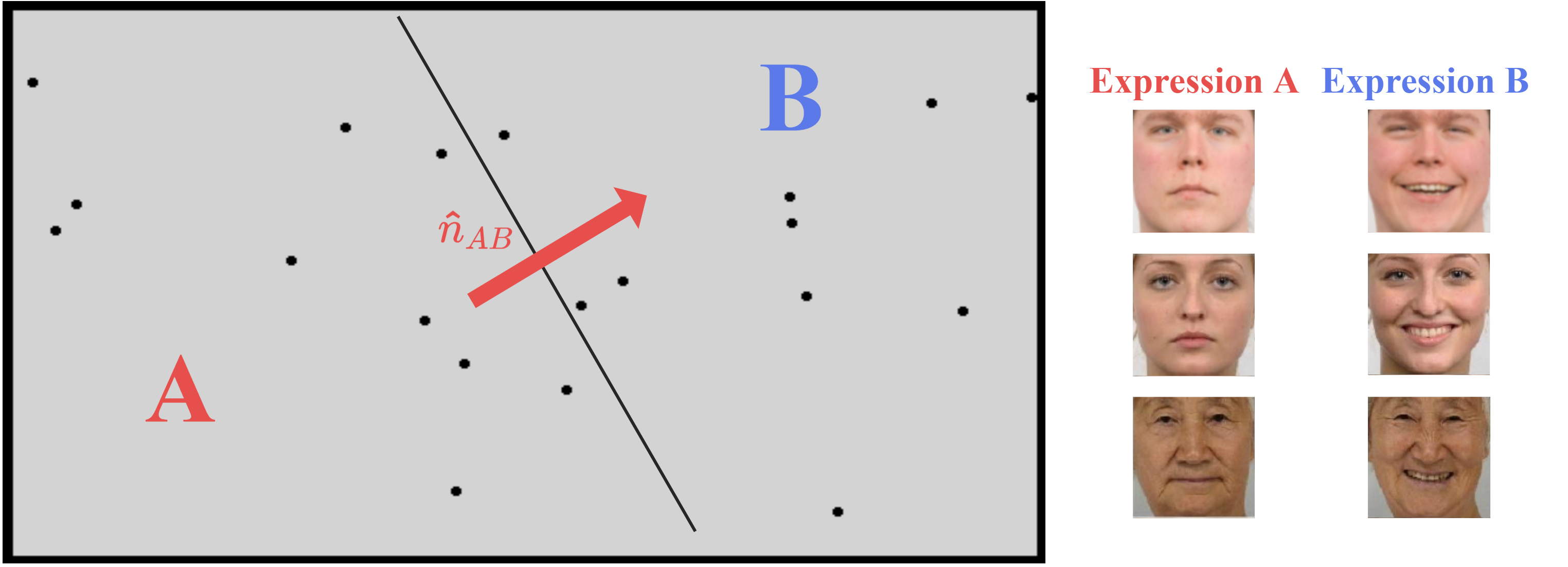}
\caption{The direction vector $\hat{n}_{AB}$ governs the transition from expression A to expression B.}
\label{fig_7_right}
\vspace{-3mm}
\end{figure}

\textbf{Step 2: Calculating facial expression direction vector.}
Following the method in Step 1, we can obtain latent vectors that share consistent identity information but exhibit different facial expressions A and B (e.g., neutral and happiness). 
Subsequently, we calculate the direction vector $n_{AB}$ (refer to Figure \ref{fig_7_right}) to achieve the transition from expression A to expression B. 
Specifically, we first obtain the sets of latent vectors for expressions A and B, denoted as $latent_A$ and $latent_B$, and assign them labels of $label_A=0$ and $label_B=1$, respectively. We then construct the mapping function $f=(1-2label_x)\vec{a} \cdot latent_x + b$, where $\vec{a}$ can be viewed as the normal vector from expression A to expression B, and the term $(1-2label_x)$ is used to control the direction\footnote[1]{When $x=A$, the term $(1-2label_A)$ in $f$ evaluates to $1$, and the calculated direction is $\vec{a}$. Conversely, when $x=B$, the term $(1-2label_B)$ in $f$ evaluates to $-1$, and the calculated direction is $-\vec{a}$, which represents the opposite direction from expression B to expression A.}. Next, we build a logistic regression model $P=\sigma(f)$ to output the predicted class label probability, which lies between $[0,1]$, and determine the normal vector $\vec{a}$ that manipulates the transition from A to B by optimizing a Binary Cross Entropy (BCE) loss. This $\vec{a}$ can be approximated as the required direction vector $n_{AB}$. For clarity, the algorithmic process for computing $\hat{n}_{AB}$ is outlined in Algorithm \ref{alg:algorithm}. 

Similarly, by adopting the aforementioned calculation algorithm, we can obtain the direction vectors between multiple facial expression states, enabling us to synthesize different types of facial expressions while preserving identity.

\begin{algorithm}[tb]
\begin{spacing}{1.0}
\caption{Calculating the direction vector $\hat{n}_{AB}$}
\label{alg:algorithm}
\textbf{Input}: Training data ($latent_x$, $label_x$), $x \in \{A, B\}$; $label_A=0$, $label_B=1$.
\begin{algorithmic}[1] %[1] enables line numbers
\STATE Initialize model parameters $\vec{a}$ and $b$ randomly
\WHILE{not converged}
\STATE $Loss = 0$
\FOR{each training sample ($latent_x$, $label_x$)}
\STATE $P=\sigma((1-2*label_x)\vec{a} \cdot latent_x + b)$
\STATE $BCELoss -= label_x*log(1-P)+(1-label_x)*log(P)$
\ENDFOR
\STATE Minimize the BCELoss by updating $\vec{a}$ and $b$ using gradient descent
\ENDWHILE
\STATE \textbf{return} $\hat{n}_{AB}=\vec{a}$
\end{algorithmic}
\textbf{Output}: The direction vector $\hat{n}_{AB}$
\end{spacing}
\end{algorithm}

%\begin{figure}[t]
%\setlength{\abovecaptionskip}{2mm}
%\centering
%\includegraphics[width=0.8\linewidth]{FER.png}
%\caption{The flowchart for facial expressions extraction based on deep learning model.}
%\label{fig_fer}
%\vspace{-3mm}
%\end{figure}

\textbf{Step 3: Multi facial expression synthesis and deep feature extraction.}

Based on the method described in Step 2, we can successfully obtain multiple direction vectors that have the ability to control different facial expressions. For any facial image, we initially apply the method mentioned in Step 1 to extract its corresponding latent vector within the latent space of StyleGAN. Subsequently, by performing a linear calculation between the latent vector and the specific direction vector that governs facial expressions, we can utilize StyleGAN's generator $G$ to synthesize a new image with the desired facial expression while preserving the identity information. The process of facial expression generation is formalized as follows:
\begin{equation}
\label{eq3}
\begin{aligned}
I^{\prime}=G_{dec}(G_{enc}(I)+\lambda\hat{n}_{AB}),
\end{aligned}
\end{equation}
where $G_{enc}$ and $G_{dec}$ denote the encoder and the decoder of $G$, respectively, $I$ represents a facial image with expression $A$, $I'$ represents the same person with expression $B$, $\hat{n}$ is the direction vector from expression $A$ to $B$, and $\lambda$ controls the degree of expression change.

Using the aforementioned facial expression synthesis method, we generate images of six basic emotions (anger, disgust, fear, sadness, happiness, surprise) from a single neutral facial image of a PD patient, capturing pre-disease expressions. These images form a control group and augment our training dataset for Parkinson's disease facial expressions. We then apply deep learning models, such as ResNet and MobileNetV3, to this enriched dataset to train discriminative models for facial expression classification and feature extraction. Our objective is to find a balance between model size and classification performance, selecting one that is both effective and suitable for mobile deployment.

%By employing the above facial expression synthesis method, we are able to generate facial expression images depicting six fundamental emotions, i.e., anger, disgust, fear, sadness, happiness, and surprise, from a single neutral facial expression image of a PD patient, capturing their expressions prior to the onset of the disease. These images serve as a control group and are used to construct an augmented training dataset of Parkinson's disease facial expressions. Subsequently, we will explore the use of various deep learning models, e.g., ResNet and MobileNetV3, on this expanded training dataset to train discriminative models. The goal is to achieve facial expression classification and extract corresponding facial features. Ultimately, we will seek a balance between model size and facial expression classification performance, selecting a deep learning model that is both performant and conducive to deployment on mobile devices.

\subsection{Multimodal Feature Fusion}
After extracting gait features and facial expression features, we propose a multimodal feature fusion strategy, namely \textbf{Hybrid Fusion}. This strategy initially processes features from both modalities through a \verb"FC" to obtain scores. These scores are appended to their respective features, creating new $m+1$-dimensional fused features. These two fused features are then input into another \verb"FC" layer, producing two-dimensional outputs. These outputs are summed to obtain the final result. This approach resembles the stacking strategy in ensemble learning, where the output class probabilities (or scores) of the primary learner serve as input features for the secondary learner.

Subsequently, we designed an end-to-end multimodal feature fusion and PD diagnosis model. During the multimodal fusion training process, we utilized pre-trained models (such as STGCN++, MobileNetV3) as feature extractors and trained only the multimodal fusion layer. In this manner, we could effectively leverage existing feature extraction capabilities and reduce training costs. Furthermore, the multimodal fusion layer will learn how to combine these two signals more effectively to extract more robust and highly discriminative diagnostic features.

\section{Experimental Results}

\subsection{Experimental Setup}
\subsubsection{Datasets Descriptions}
We use three datasets to perform the evaluations, i.e., our collected PD multimodal (PDMM) dataset of PD patients, the public facial expression dataset of normal persons Tsinghua-FED \cite{yang2020tsinghua}, and the public gait dataset of normal persons CASIA-B-HRNet \cite{fu2023gpgait}. \textbf{To the best of our knowledge, PDMM is currently the largest multimodal dataset available for in vitro diagnostic research on PD.}

The PDMM dataset was collected by our group from an ongoing population-based PD study conducted at the affiliated hospital of Nanchang University. PDMM comprises 95 PD patients (55 males and 40 females), with an average age of 62.7 and a standard deviation of 9.9. For each patient, seven images were captured using a CANON EOS 5D Mark III DSLR camera equipped with an EF 24-70mm f/2.8L II USM lens. These images represent a neutral expression and six other types of basic facial expressions (i.e., anger, disgust, fear, happiness, sadness, and surprise), along with videos recording patient behavior and gait with a duration of approximately 15 seconds. \emph{Note that the collection and use of this data have obtained informed consent from all involved PD patients. In the experiments, both the original and synthesized facial images of these PD patients were carefully censored to avoid any disclosure of personal information.}

\begin{figure}[t]
\centering
\includegraphics[width=0.92\linewidth]{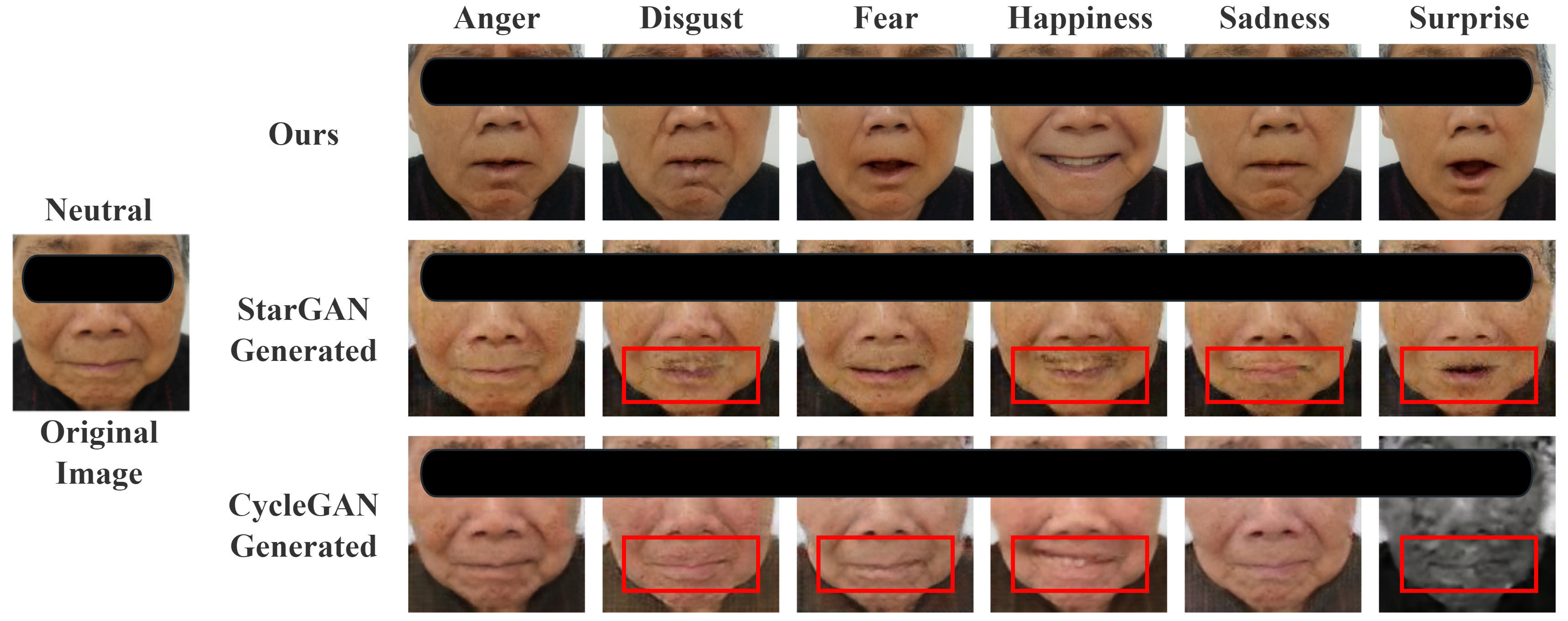}
\caption{The neutral image of a PD patient, along with the corresponding facial images encompassing the six basic emotions generated by our method, StarGAN, and CycleGAN.}
\label{fig_10}
%\vspace{-2mm}
\end{figure}

\subsubsection{Parameter Settings}

In stage 2, the values of $\lambda_{mse}$ in Eq.~(1), $\lambda_{j}$ in Eq. (2), $k$ in Eq. (2) are empirically set at $1$, $1$, $4$, respectively, as suggested in \cite{abdal2019image2stylegan}. In stage 3 of the multi-modal fusion training process, we used the pre-trained models from the first two stages (e.g., STGCN++, MobileNetV3) as feature extractors and froze their model parameters, training only the multimodal fusion layers. For this training process, we used the Adam optimizer with a learning rate of $0.001$ and conducted $100$ epochs of training to obtain the final PD diagnosis model. The proposed method was implemented using PyTorch on a server with Intel Xeon CPUs and NVIDIA Tesla T4 GPUs.

\subsection{Synthetic Facial Expression Evaluation}

In this subsection,  we evaluate the facial expression images synthesized in stage 2 of our method for six emotions: anger, disgust, fear, happiness, sadness, and surprise, and compare them with the synthesis results based on CycleGAN \cite{zhu2017unpaired} and StarGAN \cite{choi2018stargan}, as shown in Figure \ref{fig_10}. 
The results indicate that the facial expression images of PD patients generated by our method can accurately depict the six basic emotions. Compared to StarGAN and CycleGAN, which exhibit local or color distortions (as highlighted by the red box in Figure \ref{fig_10}), our method significantly outperforms in preserving facial details and maintaining image quality. These synthesis results showcase the robust image generation capability of our method using StyleGAN and also validate the effectiveness of Algorithm \ref{alg:algorithm} in generating expression direction vectors.

\begin{table}[t]
\begin{center} 
\caption{Comparison of performance of different deep learning models in facial expression classification task. Ranked in descending order according to testing accuracy.}
\label{TABLE I}
\begin{tabular}{lccc}
\hline
Model & Parameters & Train Acc.  & \textbf{Test Acc.} \\
\hline
ConvNeXt  & 106.20 MB  & 0.9822  & 0.8889 \\
\textbf{MobileNetV3}  & 5.93 MB & 0.9872  & 0.8651 \\
EfficientNetV2  & 77.85 MB & 0.9882  & 0.8651 \\
ResNet18  & 43.22 MB & 0.9901  & 0.8413 \\
MobileViT  & 3.76 MB & 0.9803  & 0.8373 \\
DenseNet121  & 27.13 MB & 0.9842  & 0.8214 \\
RegNetx\_200mf  & 9.01 MB & 0.9852  & 0.8095 \\
\hline
\end{tabular}
%\vspace{-5mm}
\end{center} 
\end{table}

\subsection{Facial Expression Feature Extraction Evaluation}
In this subsection, we utilize facial expression images of PD patients generated in Stage 2 of our method to conduct feature extraction and classification experiments on the Tsinghua-FED dataset. The objective is to evaluate the performance of various popular deep learning models, including ConvNeXt~\cite{liu2022convnet}, MobileNetV3~\cite{howard2019searching}, EfficientNetV2~\cite{tan2021efficientnetv2}, ResNet18~\cite{he2016deep}, MobileViT~\cite{wadekar2022mobilevitv3}, DenseNet121~\cite{huang2017densely} and RegNetx\_200mf~\cite{radosavovic2020designing}, on facial expression feature extraction and classification tasks. We adopt a 4:1 partitioning strategy, dividing the facial expression dataset into training and testing sets, and report the training and testing accuracies of different deep learning models in Table \ref{TABLE I}. It is observed that all involved deep learning models achieve a testing accuracy exceeding 80\%, and some lightweight models, such as MobileNetV3 and MobileViT, achieve performance comparable to large parameter models. Considering the intended application of our method in mobile device scenarios for remote healthcare, lightweight models with higher deployability are prioritized for facial feature extraction. Therefore, we choose MobileNetV3, which achieves a good balance between accuracy and efficiency, as the facial expression feature extractor in Stage 2.

\subsection{PD Diagnosis Evaluation}
%In this subsection, we initially evaluate the performance of the two different feature fusion strategies proposed in stage 3 for the diagnosis of PD. Specifically, we follow the evaluation protocol in \cite{huang2022facial,zhou2024early}, and divide the PDMM dataset into 5 folds, with 4 folds (76 PD patients) utilized for training and the remaining 1 fold (19 PD patients) used for testing. To further enhance the testinig set, we incorporate facial expression images of 47 subjects aged above 60 from the Tsinghua-FED dataset, along with 47 gait video clips from the CASIA-B-HRNet dataset, forming 47 non-PD control groups. 
%Therefore, a total of 66 subjects are used for PD diagnosis during the testing phase. As shown in the table, the two strategies exhibit comparable performance. Additionally, when using a weaker feature extractor like MobileViT, the second hybrid fusion strategy slightly outperforms the first concatenation strategy, benefiting from the stacking concept in ensemble learning.

In this subsection, we compare our method using the hybrid fusion strategy with three latest representative PD diagnosis methods: co-occurrence matrix (GLCM) combined with support vector machine (SVM), dubbed GLCM+SVM \cite{hou2022visual}, the data-driven unimodal facial expression PD diagnosis methods StarGAN+ResNet \cite{huang2022facial} and FEPD \cite{zhou2024early}, and three popular deep learning models, namely DeiT \cite{touvron2021training}, ConvNeXt \cite{liu2022convnet}, and EfficientNetV2 \cite{tan2021efficientnetv2}. Specifically, we follow the evaluation protocol in \cite{huang2022facial,zhou2024early}, and divide the PDMM dataset into 5 folds, with 4 folds (76 PD patients) utilized for training and the remaining 1 fold (19 PD patients) used for testing. To further enhance the testinig set, we incorporate facial expression images of 47 subjects aged above 60 from the Tsinghua-FED dataset, along with 47 gait video clips from the CASIA-B-HRNet dataset, forming 47 non-PD control groups. 
Therefore, a total of 66 subjects are used for PD diagnosis during the testing phase. All the three deep learning models mentioned above are fine-tuned using the PDMM training dataset. 

%Furthermore, we compare our method using the hybrid fusion strategy with three latest representative PD diagnosis methods: co-occurrence matrix (GLCM) combined with support vector machine (SVM), dubbed GLCM+SVM, the data-driven unimodal facial expression PD diagnosis methods StarGAN+ResNet and FEPD, and three popular deep learning models, namely DeiT, ConvNeXt , EfficientNetV2, and FaceQNet \cite{hernandez2019faceqnet}.

%Furthermore, we compare our method using the hybrid fusion strategy with three latest representative PD diagnosis methods: co-occurrence matrix (GLCM) combined with support vector machine (SVM), dubbed GLCM+SVM, the data-driven unimodal facial expression PD diagnosis methods StarGAN+ResNet and FEPD, and two popular deep learning models, namely ConvNeXt and EfficientNetV2.

\begin{table}[t]
\begin{center} 
\caption{Comparison between our method and the other methods w.r.t. PD diagnosis accuracy.\label{tab:diagnosis_accuracy}}
\begin{tabular}{lc}
\hline
Diagnosis Method & Accuracy \\
\hline
%FaceQNet \cite{hernandez2019faceqnet} & 0.7603 \\
EfficientNetV2 \cite{tan2021efficientnetv2} & 0.7125\\
DeiT-small \cite{touvron2021training}& 0.7193 \\
ConvNeXt-Tiny \cite{liu2022convnet}& 0.7336\\
GLCM+SVM \cite{hou2022visual} & 0.4660 \\
StarGAN+ResNet18 \cite{huang2022facial} & 0.9543 \\
FEPD \cite{zhou2024early}  & 0.9755  \\ \hline
\textbf{Our proposed method} & \textbf{1}\\
\hline
\end{tabular}
\end{center} 
\vspace{-3mm}
\end{table}

The diagnostic accuracies of our method and the other comparison methods are presented in Table \ref{tab:diagnosis_accuracy}. It can be observed that our method achieves the highest diagnostic accuracy for PD, outperforming the unimodal methods StarGAN+ResNet and FEPD, and significantly surpassing the traditional GLCM+SVM PD diagnosis method as well as the other deep learning models. The superiority of our method can be attributed to its comprehensive analysis by fusing behavioral gait and facial expression features, extracting gait features with a powerful STGCN++, and proposing a multi-class facial expression generation scheme based on StyleGAN to facilitate the extraction of expressive features. The effective fusion of these two types of features mitigates the misdiagnosis issues associated with unimodal methods in PD diagnosis, thereby enhancing the accuracy of PD diagnosis. Additionally, the conventional machine learning-based GLCM+SVM performs poorly compared to the three deep learning methods and is far inferior to our method in terms of PD diagnosis accuracy, demonstrating the good representation learning capability of deep neural networks.

\subsection{Ablation Study}
\begin{table}[t]
\centering
\caption{PD diagnosis results using unimodal features of facial expressions or behavioral gait.}
\label{tab:model_comparison}
 \begin{tabular}{llc}
\hline
& Model & \textbf{Test Acc.} \\
\hline
\multirow{4}{*}{\parbox{3cm}{Unimodal\\ Facial Expression\\ Diagnosis}} & MobileNetV3 & 0.9692 \\ 
& ConvNeXt  & 0.9538 \\
& EfficientNetV2 & 0.9538 \\
& ResNet18 & 0.9230 \\
%& MobileViT & 0.8615 \\
\hline
%& Model & \textbf{Test Acc.} \\
%\hline
\multirow{2}{*}{\parbox{3cm}{Unimodal Behavior Gait Diagnosis}} & STGCN++ & 0.9692 \\
& STGCN & 0.9385 \\
\hline
\end{tabular}

\vspace{-1mm}
\end{table}

In this subsection, we further explore the performance of our method when using only unimodal facial expression or gait features for PD diagnosis. For unimodal PD diagnosis based on facial expressions, we adopt not only the default MobileNetV3 as the feature extractor but also experiment with other deep model extractors such as ConvNeXt, EfficientNetV2, and ResNet18, aiming to assess their performance in PD diagnosis. Due to the limited availability of skeletal feature extraction models for gait-based PD diagnosis, we use only the default STGCN++ and its previous version, STGCN, for comparative analysis. The specific results are shown in Table \ref{tab:model_comparison}. It is evident that for facial expression-based PD diagnosis, the performance of different feature extractors in PD diagnosis is closely related to their respective models' performance in classification tasks. The ranking of PD diagnosis performance is broadly consistent with the facial expression classification results shown in Table \ref{TABLE I}, indicating the importance of selecting models based on facial expression classification performance for PD diagnosis. For gait-based PD diagnosis, STGCN++ slightly outperforms STGCN. Notably, both unimodal facial expression-based PD diagnosis (using MobileNetV3) and unimodal gait-based PD diagnosis (using STGCN++) exhibit lower performance in PD diagnosis compared to our multimodal method, which combines facial expression and gait features. This not only validates the rationality of fusing these two features for PD diagnosis but also demonstrates the effectiveness of our proposed feature fusion strategy.

\section{Conclusion}

%This paper has proposed a novel in vitro diagnostic method for PD, which is the first attempt to fuse multi-modal data from facial expressions and behavioral gait, and is expected to become an important advancement in this field. On one hand, we propose a multi-class facial expression generation scheme based on StyleGAN, which aims to synthesize pre-morbid facial expression data of PD patients, facilitating the extraction of expressive features. On the other hand, we employ the lightweight STGCN++ for gait feature extraction. Furthermore, we introduce two different modes of feature fusion schemes to effectively integrate facial expression features with gait features, thereby enhancing the accuracy of PD diagnosis. Empirical studies have validated the superiority of the proposed method in PD diagnosis.

%This paper introduces a novel in vitro diagnostic approach for Parkinson's disease (PD), marking the first attempt to integrate multi-modal data from facial expressions and behavioral gait. On one hand, we propose a StyleGAN-based multi-class facial expression generation framework to synthesize pre-morbid facial data of PD patients, aiding in the extraction of expressive features. On the other hand, we leverage the lightweight STGCN++ for gait feature extraction. Moreover, we devise a hybrid feature fusion strategy to seamlessly combine facial and gait features, thereby boosting the accuracy of PD diagnosis. Empirical studies have demonstrated the superiority of our proposed method in PD diagnosis.

This paper presents a novel in vitro diagnostic method for Parkinson’s Disease (PD), pioneering the integration of multimodal facial expression and gait behavior data. We introduce (1) a StyleGAN-based framework for synthesizing pre-morbid PD facial expressions to enhance expressive feature extraction, and (2) a lightweight STGCN++ model for gait analysis. A hybrid fusion strategy effectively combines facial and gait features, significantly improving PD diagnostic accuracy. Experimental results validate the superiority of our approach in PD diagnosis.

\section{Acknowledgement}

This work is supported in part by Natural Science Foundation of China (62466036, 62271239), by Natural Science Foundation of Jiangxi Province (20232BAB212025, 20232BAB206065), by High-level and Urgently Needed Overseas Talent Programs of Jiangxi Province (20232BCJ25024), and by Jiangxi Double Thousand Plan (JXSQ2023201022).

%\small
\bibliographystyle{apacite}

\setlength{\bibleftmargin}{.125in}
\setlength{\bibindent}{-\bibleftmargin}

\bibliography{cogsci}

\end{document}